\documentclass{INTERSPEECH2023}

\interspeechcameraready

\title{Human Transcription Quality Improvement}
\name{Jian Gao$^1$, Hanbo Sun$^1$, Cheng Cao$^1$, Zheng Du$^1$}
\address{
  $^1$Amazon, USA
}
\email{gajian@amazon.com, sunhanbo@amazon.com, chengcao@amazon.com, zhengdu@amazon.com}

\begin{document}

\maketitle
 
\begin{abstract}
% High quality transcription data is crucial for training automatic speech recognition (ASR) systems. However, existing data collection methods are expensive to researchers, while the quality of crowdsourced transcription is often low. In this paper, we propose a novel human-in-the-loop framework of transcription data collection. We introduce two mechanisms to improve transcription quality: confidence estimation based reprocessing at the labeling stage, and automatic word error correction at post-labeling stage. Moreover, we collect and release LibriCrowd - a large-scale crowdsourced dataset of audio transcription on 100 hours of English speech. Experiment shows our method can reduce transcription WER by over 50\%. We further investigate the impact of transcription error on ASR model performance and found a strong correlation, where the quality improvement provides over 10\% relative WER reduction. Finally, we release a dataset of crowdsourced transcription and the code.
High quality transcription data is crucial for training automatic speech recognition (ASR) systems. However, the existing industry-level data collection pipelines are expensive to researchers, while the quality of crowdsourced transcription is low. In this paper, we propose a reliable method to collect speech transcriptions. We introduce two mechanisms to improve transcription quality: confidence estimation based reprocessing at labeling stage, and automatic word error correction at post-labeling stage. We collect and release LibriCrowd - a large-scale crowdsourced dataset of audio transcriptions on 100 hours of English speech. Experiment shows the Transcription WER is reduced by over 50\%. We further investigate the impact of transcription error on ASR model performance and found a strong correlation. The transcription quality improvement provides over 10\% relative WER reduction for ASR models. We release the dataset and code to benefit the research community.
\end{abstract}

\section{Introduction}

Speech recognition systems have made significant progress during the recent years. The research community is actively developing algorithms for automatic speech recognition (ASR), and many models are at par with or even surpass humans \cite{8049322}. However, less work has been published on the technology of manual speech recognition (MSR), though high quality human speech transcription is crucial for all spoken language related research.

The development of speech corpus produced by MSR follows the design of ASR system. Two decades ago, hybrid systems \cite{10.5555/562393} were popular, in which Hidden Markov Models (HMMs) learn the transition between phones, and Recurrent Neural Networks (RNNs) perform localized classifications. Speech corpus like TIMIT \cite{TIMIT1993} provides manually time-aligned phonetic and word transcriptions to support phone-level training and evaluation. Due to the high cost of  hand-aligning each word with its corresponding audio frames, people use forced alignment \cite{mcauliffe17_interspeech} to automatically generate word aligned datasets. Later, CTC \cite{Graves:06icml} was proposed to train ASR models on unsegmented data by aggregating over all possible alignments, and word-aligned datasets are no longer necessary.  

In the 2010s, ASR systems were greatly improved and became hungry for data. TIMIT was found too small to train the prediction network in RNN-T \cite{DBLP:journals/corr/abs-1211-3711}. This trend is similar as in the vision and language domains. GPT-3 \cite{NEURIPS2020_1457c0d6} shows language model (LM) performance scales as a power-law of dataset size, model size, and computation power. Compared with text and image datasets, the scale of speech corpus is lagging far behind and many tasks are short of supervised data. To solve this problem, many recent works \cite{NEURIPS2020_92d1e1eb,9688253,9585401} move to unsupervised or self-supervised learning that use mostly unlabeled audio plus a small amount of transcribed speech. For example, with 10-hour labeled speech , WavLM \cite{9814838} achieves similar word error rate (WER) as many fully supervised models \cite{9053896,gulati20_interspeech,han20_interspeech} trained on the entire 960-hour LibriSpeech corpus \cite{7178964}. Nevertheless, the data problem cannot be bypassed as labeled data is crucial for real-world applications in which the ASR model performance is often much worse than published benchmark results. To bridge this gap, the investment on quality controlled data collection is as important as improving ASR algorithms.

Today's industry-level data annotation pipelines \cite{DBLP:journals/corr/abs-2109-01164} are often expensive and inconvenient for researchers, even though they hire professional transcribers. The standard transcription cost in U.S. is around \$90 per speech hour. To save cost, some data providers use synthetic data or ASR-assisted pre-labeling to replace humans, which can introduce unwanted bias \cite{levit2017dont}. 

Crowdsourcing has been prevalent to produce the majority of labeled data in many computer vision tasks, replacing professional annotators.  Platforms like Microsoft UHRS, Amazon MTurk, Appen, and Scale AI distribute annotation tasks to a large group of people, which is often cheap and efficient. However, for complex tasks like speech recognition, crowdsourced annotators who lack certified training are likely to provide low-quality annotations.

Thus far, few research has quantitatively analyzed the impact of human transcription error on ASR models. \cite{1325949} claims HMM-based ASR systems are robust to mislabeled transcriptions as Gaussian Mixture Models learn very little from incorrect labels. However, \cite{DBLP:journals/corr/abs-2112-00350} has conflicting observations on RNN-T based ASR systems. It shows the negative impact brought by transcription defects, especially the deletion error, cannot be recovered despite increasing model size and data size. Nevertheless, the transcription errors found in both \cite{1325949} and \cite{DBLP:journals/corr/abs-2112-00350} are from the simulation rather than by real human transcribers. \cite{DBLP:journals/corr/abs-2112-00350} uses soundex \cite{odell1956profit} and bi-gram language model to simulate substitution and insertion errors. The simulated error patterns cannot well represent the mistakes made by real human. 

\begin{figure*}[ht]
  \centering
  \includegraphics[width=\linewidth]{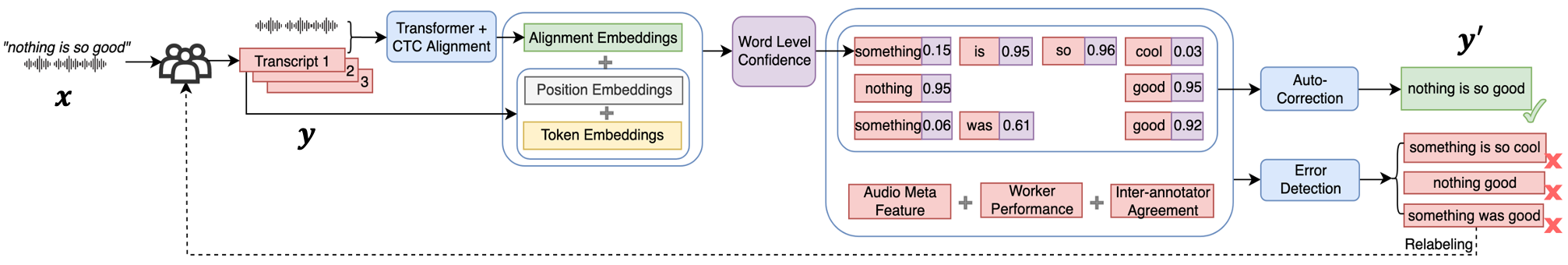}
  \caption{The workflow of proposed ML-in-the-loop human transcription pipeline, where multiple crowdsourced workers transcribe one audio at a time. The audio-text alignment module generates the alignment embeddings, which are the input of a transformer to predict word-level confidence scores. The aligned transcripts among multiple transcripts and those confidence scores are used for auto-correction. Along with other meta features, they are fed to a model to predict sentence level likelihood of transcription error.}
  \label{fig:pipeline}
  \vspace{-3mm}
\end{figure*}

In this paper, we propose a ML-in-the-loop data collection mechanism for high-quality speech transcription. It includes two schemes of quality improvement. First, during annotator's labeling, confidence estimation modules (CEMs) identify low-quality transcriptions for relabeling. Next, after annotator's labeling, error correction models (ECMs) generate the final output with reduced Transcription WER (TWER). Through experiments we find the two stages together provide a 50\% TWER reduction. To analyze the transcription error’s impact on training ASR models, we train Wav2Vec2 \cite{NEURIPS2020_92d1e1eb} and WavLM \cite{9814838} on noisy transcription labels at controlled TWER. We find a strong correlation between TWER and WER of downstream ASR models. The quality-improved training data can provide over 10\% WER reduction. In summary, the contributions of this paper are:

\begin{enumerate}
\item We propose a new ML-in-the-loop data collection mechanism to produce high-quality transcriptions for ASR model training. Experiment shows it can significantly improve the human transcription quality. %The code is released so that individual researchers can easily create their own speech dataset. 
\item We analyze the impact of transcription error on training ASR models, with incorrectly transcribed data by real human. We find a strong correlation between TWER and the WER of downstream ASR models. %It shows the importance of collecting high quality transcription labels.
\item We collect and release LibriCrowd\footnote{\url{https://github.com/GenerateAI/LibriCrowd}}, a large-scale public dataset of crowdsourced audio transcriptions. It contains 100 hours of English speech transcribed by 4433 human transcribers. The labeling cost is only \$6 per speech hour. We believe LibriCrowd can benefit the research community to develop new MSR error correction models and design robust ASR models trained on noisy data. 
\end{enumerate}

\section{Data Collection}

This section introduces the new data collection mechanism we proposed. We collect speech transcription data via crowdsourcing from Amazon MTurk. Compared with the existing industry-level data pipelines \cite{DBLP:journals/corr/abs-2109-01164,levit2017dont,DBLP:conf/nips/PavlichenkoSU21}, our data collection approach does not require expensive worker training, machine pre-labeling, human auditing, and annotator behavior monitoring. The transcription quality is controlled by the proposed confidence estimation modules (CEMs) at the labeling stage and error correction models (ECMs) at the post-labeling stage.

\subsection{Data Source}

% In this paper we use a large-scale speech corpus as the data source
In this work, we start from raw audios and collect human transcriptions to build the speech corpus. Our setting differs from many existing speech corpora \cite{TIMIT1993,7178964,ardila-EtAl:2020:LREC} that were built on text-to-speech reading, since prepared scripts are rare in real-world speech applications. The audio files are collected from the LibriVox project, in which ground truth reference is available for us to measure the crowdsourced transcription quality. 

% Many existing speech corpora \cite{TIMIT1993,7178964,ardila-EtAl:2020:LREC} are built upon text-to-speech reading, but prepared scripts are rare in real-world speech. In this work, we start from raw audio and collect human transcriptions to build the speech corpus. This setting naturally fits in real-world speech applications. Meanwhile, we need the "ground truth" reference to measure the human transcription quality, but it is usually difficult to have it for unscripted speech. Therefore, we use publicly available scripted speech data as the data source. 
% Therefore, we use publicly available scripted speech data and collect human transcription through crowdsoucing.  %The label quality is measured by comparing it with the existing ground truth reference.

% such as spontaneous conversation or voice search query due to the highly ambiguous nature. 
%  labels by crowdsourcing. 

% We create LibriCrowd, 

%  a speech corpus of human transcription as the data source.

\subsection{Transcription Task Design}

% Though audio transcription looks natural to people, proper task design can improve quality as well as reduce the cost. 

\noindent \textbf{User Interface and Instruction}: The UI is designed based on MTurk’s audio transcription template\footnote{\url{https://github.com/GenerateAI/TransAudioUI}}. Each task contains speech recordings loaded from external sources and blank fields for human annotators to provide transcription inputs. We increase the space between the text field and submit button to avoid incomplete submissions. To align with the format of ground truth transcription, we set instructions such as ignoring punctuation marks, transcribing digits as words, etc. Annotators are allowed to put a question mark at the positions where they are uncertain, instead of random guess. 

\noindent \textbf{Reward per Utterance}: The trade-off between labeling cost, quality and efficiency has been studied in \cite{5494979,novotney-callison-burch-2010-cheap,5947474}. In many real-world applications the cost per utterance ranges from 0.5 to 5 cents, and it will mainly affect the efficiency rather than quality. Hence in this study we set reward at \$0.01 per task, each of which contains five utterances.  

\noindent \textbf{Worker Selection}: We do not set entrance exam and training requirement, unlike many existing industry-level transcription pipelines. All workers with the approval rate $\geq 95\%$ are qualified. The purpose is to reduce cost and increase diversity. Later the confidence estimation models we build will detect those low-quality submissions and send to a different worker for relabeling. We use the model results to filter out malicious or careless workers.

\noindent \textbf{Number of Workers per Utterance}: Each audio recording is assigned to five different workers. The inter annotator agreement will be an important signal for our confidence estimation afterwards. Our error correction models aggregate all transcripts to generate the final transcription.

\section{Quality Improvement}

We develop confidence estimation model (CEM) and error correction model (ECM) for quality improvement as soon as we collect every human transcription, as shown in Figure~\ref{fig:pipeline}. Transcribers listen to the audio $\bm{x}$ and produce raw transcriptions $\bm{y}$. CEM detects error for every word in $\bm{y}$ and sends low-quality utterances for relabeling. ECM automatically corrects error words in $\bm{y}$ and generates transcriptions $\bm{y'}$ as the final output. 

% Given an audio sequence $\bm{x}$ and noisy transcription $\bm{y}$, CEMs detect errors in $\bm{y}$ and send it for relabeling; ECMs generate clean transcription $\bm{y'}$ by correcting the error words. 

\subsection{Confidence Estimation}

Confidence scores are calculated at word and utterance level to assess the MSR quality. Various factors are considered, such as spell and grammar error, audio-text mismatch, transcriber’s recent performance record, speech length and complexity. %An ensemble model is built on top of them to make predictions. 

% We first build a word-level 
CEM takes paired audio-text sequence $(\bm{x},\bm{y})$ as input, and generates a confidence score $c_i$ for each word $y_i \in \bm{y}$. We extract speech representations by pre-trained Wav2Vec2 \cite{NEURIPS2020_92d1e1eb} on LV-60k, and then use CTC-Segmentation \cite{10.1007/978-3-030-60276-5_27} to align each word with its corresponding audio segment. Suppose $\bm{x}_{u_i:v_i}$ is the audio segment of the $i^{th}$ word $y_i$, the alignment score $s_i$ is calculated based on the emission probability $P(y_i|\bm{x}_{u_i:v_i})$ and CTC loss \cite{Graves:06icml} given its bidirectional context. 
\begin{equation}
\begin{aligned}
     \log s_i(\bm{x},\bm{y}) = \log P(y_i | \bm{x}, \bm{y}_{-i}) = \log P(y_i|\bm{x}_{u_i:v_i}) \\
     - \mathcal{L}_{CTC}(\bm{x}_{1:u_{i-1}},\bm{y}_{1:i-1}) -  \mathcal{L}_{CTC}(\bm{x}_{v_{i+1}:S},\bm{y}_{i+1:T})   
\end{aligned}
\end{equation}
where $\bm{x}_{u_i:v_i} = (x_{u_i},\cdots,x_{v_i})$, $\bm{y}_{1:i}=(y_1,\cdots,y_i)$, $\bm{y}_{-i}=(y_1,\cdots,y_{i-1}, y_{i+1},\cdots,y_T)$. $S = |\bm{x}|$ is the audio length, and $T = |\bm{y}|$ is the transcription length.

To further detect those errors from either syntactic or semantic perspective, we modify a BERT-like error detection model ELECTRA \cite{9688175} by adding the alignment score $s_i$ as the embedding. The discriminator of ELECTRA is fine-tuned on noisy transcriptions with real human errors. It is aligned with the ground truth reference and each word is labeled as correct or incorrect. The model outputs a confidence score $c_i$ for each word $y_i$. 

Then we train a model for utterance-level error detection. It predicts the expected number of word errors in an utterance based on the word-level CEM as well as meta features from three sources: (1) utterance complexity features such as audio duration, transcription length, and signal-to-noise ratio (SNR) estimated by the WADA-SNR algorithm \cite{kim08e_interspeech}; (2) worker performance features such as task spending time, recent task accept rate; (3) inter annotator agreement, i.e., the edit distance from one transcription to other transcriptions for the same audio recording. The model is built upon gradient boosting trees and implemented by LightGBM \cite{NIPS2017_6449f44a}. The model provides the real-time feedback to transcribers by rejecting those unqualified responses and republishing the tasks.

\subsection{Error Correction}

% It’s popular to use language models like BERT \cite{devlin-etal-2019-bert} and BART \cite{lewis-etal-2020-bart} for text error correction. 
Language models like BERT \cite{devlin-etal-2019-bert} and BART \cite{lewis-etal-2020-bart} have been widely used for text error correction. It is also well-studied to refine ASR outputs by second-pass rescoring on n-best hypotheses \cite{9747118}. For MSR error correction, ECM can do word-level aggregation instead of sentence-level reranking in ASR error correction. This is because MSR outputs have independent sources while the n-best ASR hypotheses are generated by beam search decoding from the same acoustic model. However, unlike ASR system, human transcription does not have a confidence score generated by the first-pass decoder. 

Therefore, we use the confidence score estimated by the word-level CEM. It measures the audio-text alignment as well as semantic consistency. Meanwhile, we perform text-text alignment between multiple human transcriptions of the same audio recording. The alignment is conducted by minimizing the edit distance between two text sequences using the Needleman–Wunsch Algorithm \cite{NEEDLEMAN1970443}. The correction is based on a simple voting scheme at each aligned position. 

For a given utterance, suppose $w=y_i$ is the $i^{th}$ word, $N(w,i)$ is the occurrence of word $w$ at position $i$, $N$ is the number of candidate transcriptions, and $c_i(w)$ is the confidence score. Then the overall scoring of word $w$ is 
\begin{equation}
\begin{aligned}
    & s_i(w) = \alpha N(w,i)/N + (1-\alpha)c_i(w) \\
    & w^* = \arg\max_{w} s_i(w) 
\end{aligned}
\end{equation}
where $w^*$ is the selected word at position $i$. The trade-off parameter $\alpha$ between word frequency and confidence score is tuned on the training set. Note that when $\alpha=1$, our algorithm degrades to ROVER \cite{659110}. Figure \ref{fig:pipeline} includes an example for ECM.

\section{Experiment}

In this section, we evaluate the quality of MSR and analyze the impact of transcription error on ASR model training. MSR output and ASR output are compared with the ground truth reference to calculate TWER and WER, respectively.

\subsection{Experiment Setup}

We experiment on speech recordings with ground truth scripts to measure the quality of human transcription. Note that the ground truth is not available in most real scenarios. A surrogate is the consensus annotation of multiple expert transcribers. 

For data source, we use audio recordings from LibriVox to create LibriCrowd. LibriVox\footnote{\url{https://librivox.org/}} is a free public project that contains thousands of audio books read by a group of worldwide volunteers. The ground truth is directly from the text in the audio book. LibriCrowd contains 100 hours of English speech in three categories: (1) 70 hours audio recordings from higher-WER speakers (\textit{train-other-10h}, \textit{train-other-60h}), (2) 10 hours Limited Resource Training Set of Libri-Light \cite{9052942} (\textit{train-mixed-10h}), (3) Dev and Test Sets of LibriSpeech (20 hours). We release LibriCrowd to the research community under CC-BY-4.0 license. The statistics of LibriCrowd is summarized in Table \ref{tab:1}.

% Table 1: LibriCrowd Dataset Statistics and Raw Transcription Word Error Rate (TWER) without Quality Control
\begin{table}[]
\caption{LibriCrowd dataset statistics and raw TWER without quality improvement}
\vspace{-1mm}
\label{tab:1}
\resizebox{\columnwidth}{!}{%
\begin{tabular}{c|ccccc}
\hline
\textbf{Subset} & \textbf{\# Utterances} & \textbf{speech hours} & \textbf{\# Workers} & \textbf{\# Responses} & \textbf{TWER (\%)} \\ \hline
train-other-10h  & 3165  & 10.0  & 1258 & 18673  & 15.50 \\ \hline
train-other-60h  & 17816 & 60.0  & 1136 & 20187  & 7.52  \\ \hline
train-mixed-10h  & 2763  & 9.8   & 616  & 14231  & 5.97  \\ \hline
dev-clean        & 2703  & 5.4   & 523  & 13994  & 6.10  \\ \hline
test-clean       & 2620  & 5.4   & 527  & 13587  & 8.23  \\ \hline
dev-other        & 2864  & 5.3   & 620  & 15235  & 12.69 \\ \hline
test-other       & 2939  & 5.1   & 989  & 15950  & 16.61 \\ \hline
all              & 34870 & 101.0 & 4433 & 111857 & 10.91 \\ \hline
\end{tabular}%
}
\end{table}

The speech recordings in LibriCrowd are transcribed in a cold-start setting. For the first 10 hours of speech (\textit{train-other-10h}), human adjudicators are hired to verify the submitted transcriptions. They listen to the audio and republish their transcriptions to replace the low-quality ones. This data is used to train CEM and ECM.

\subsection{MSR Transcription Quality Analysis}

Table \ref{tab:2} lists the performance of word-level CEM. Compared to the ELECTRA baseline, having the alignment embedding can improve F1 score by 9\% on LibriCrowd. The audio-text alignment serves as a surrogate “ASR confidence” which provides a reliable confidence estimation for human transcription.

% Table 2: Word-Level Transcription Error Detection Model Performance. 
\begin{table}
\centering
\caption{Word-level confidence estimation on LibriCrowd}
\vspace{-1mm}
\label{tab:2}
\resizebox{0.35\textwidth}{!}{%
\begin{tabular}{l|rrr}
\hline
                         & \multicolumn{1}{c}{Precision} & \multicolumn{1}{c}{Recall} & \multicolumn{1}{c}{F1} \\ \hline
ELECTRA                  & 0.6668 & 0.8316 & 0.7401 \\ \hline
ELECTRA + Alignment      & 0.7590 & 0.8615 & 0.8070 \\ \hline
% Utterance-Level Ensemble &        &        &        \\ \hline
\end{tabular}%
}\vspace{-4mm}
\end{table}

Table \ref{tab:3} shows the MSR transcription quality before and after applying the proposed quality improvement approach. In summary, CEM-based relabeling can reduce 22\% TWER (10.91 → 8.48), and ECM further reduces 42\% TWER (8.48 → 4.94). Together, our approach reduce 55\% TWER from raw transcription (10.91 → 4.94). Note that our method is much better and cheaper than CrowdSpeech \cite{DBLP:conf/nips/PavlichenkoSU21} (4.94 v.s 7.84) in which each recording was transcribed by seven workers.
% Table 3: Comparison of the proposed quality control models with baselines and the oracle performance. MSR transcription quality is evaluated by TWER against the ground truth transcription (lower value is better). 
\begin{table}[]
\caption{Comparison of proposed quality improvement methods with baselines. Quality is evaluated by TWER (\%) against ground truth. A lower value means higher quality.}
\vspace{-1mm}
\label{tab:3}
\resizebox{\columnwidth}{!}{%
\begin{tabular}{l|cccccc}
\hline
\textbf{} &
  \multicolumn{1}{l}{train-mixed-10h} &
  \multicolumn{1}{l}{dev-clean} &
  \multicolumn{1}{l}{test-clean} &
  \multicolumn{1}{l}{dev-other} &
  \multicolumn{1}{l}{test-other} &
  \multicolumn{1}{l}{average} \\ \hline
Raw Transcription        & 5.97 & 6.10 & 8.23 & 12.69 & 16.61 & 10.91 \\ \hline
CEM + Random                 & 5.00 & 4.94 & 5.77 & 10.20 & 13.01 & 8.48  \\ \hline
CEM + Longest                & 5.12 & 5.83 & 6.30 & 10.20 & 12.42 & 8.69  \\ \hline
CEM + Best Worker            & 4.60 & 4.72 & 5.17 & 9.51  & 11.30 & 7.68  \\ \hline
CEM + RescoreBERT            & 4.11 & 4.58 & 5.96 & 9.99  & 11.14 & 7.16  \\ \hline
CEM + ECM &
  \textbf{2.99} &
  \textbf{2.76} &
  \textbf{3.05} &
  \textbf{6.45} &
  \textbf{7.51} &
  \textbf{4.94} \\ \hline
CEM + Oracle                 & 2.05 & 1.69 & 1.88 & 4.13  & 5.00  & 3.18  \\ \hline
CrowdSpeech + ROVER  & N/A  & 6.76 & 7.29 & 13.19 & 13.41 & 10.16 \\ \hline
CrowdSpeech + T5     & N/A  & N/A  & 5.22 & N/A   & 10.46 & 7.84  \\ \hline
CrowdSpeech + Oracle & N/A  & 3.81 & 4.32 & 8.26  & 8.50  & 6.22  \\ \hline
\end{tabular}%
}
\end{table}

At the labeling stage, the average relabeling rate is 5.4\%. In those rejected transcriptions, deletion error dominates the TWER, while in the accepted responses, substitution error is prevalent (Table \ref{tab:4}). We find some rejected samples are empty or incomplete. Through the email communication, some transcribers mentioned they mis-clicked the submit button before finishing the task. Then we slightly increased the gap between the submit button and the transcription text box. The reduced deletion error can reflect the TWER gap between those rejected transcriptions in \textit{test-clean} and \textit{dev-clean}. This finding shows a good UI design can improve human transcription quality. 

When we published the tasks we followed in the order of \textit{train-other-10h}, \textit{test-other}, \textit{dev-other}, \textit{test-clean}, \textit{dev-clean}, and \textit{train-mixed-10h}. Through that process we observed the trend of improving worker performance and decreasing worker number. It shows CEM-based quality assessment can effectively eliminate invalid task submissions and help filter out poorly-performed workers. Meanwhile, the frequency of substitution error reduced (17.56 → 5.32), thanks to the alignment embedding that detects audio-text mismatch. 

% Table 4: Error Type Analysis in Raw MSR Transcriptions
\begin{table}[]
\caption{Error types in raw MSR transcription}
\vspace{-1mm}
\label{tab:4}
\resizebox{\columnwidth}{!}{%
\begin{tabular}{l|cccccc}
\hline
\multicolumn{1}{c|}{Accepted} &
  \multicolumn{1}{l}{train-mixed-10h} &
  \multicolumn{1}{l}{dev-clean} &
  \multicolumn{1}{l}{test-clean} &
  \multicolumn{1}{l}{dev-other} &
  \multicolumn{1}{l}{test-other} &
  \multicolumn{1}{l}{average} \\ \hline
Length (\# word) & 35    & 20    & 20    & 18    & 18    & 22    \\ \hline
Deletion (\%)    & 0.87  & 0.91  & 1.17  & 1.67  & 2.77  & 1.48  \\ \hline
Insertion (\%)   & 0.44  & 0.42  & 0.48  & 0.95  & 1.33  & 0.72  \\ \hline
Substitution (\%)& 3.58  & 3.68  & 4.12  & 6.90  & 8.34  & 5.32  \\ \hline
TWER (\%)        & 4.89  & 5.01  & 5.78  & 9.52  & 12.43 & 7.53  \\ \hline
\multicolumn{1}{c|}{Rejected} &
  \multicolumn{1}{l}{} &
  \multicolumn{1}{l}{} &
  \multicolumn{1}{l}{} &
  \multicolumn{1}{l}{} &
  \multicolumn{1}{l}{} &
  \multicolumn{1}{l}{} \\ \hline
Length (\# word) & 17    & 12    & 9     & 9     & 9     & 11    \\ \hline
Deletion (\%)    & 44.72 & 33.46 & 57.22 & 48.94 & 49.31 & 46.73 \\ \hline
Insertion (\%)   & 1.74  & 2.71  & 2.43  & 3.54  & 3.91  & 2.87  \\ \hline
Substitution (\%)& 15.40 & 17.49 & 15.12 & 20.47 & 19.34 & 17.56 \\ \hline
TWER (\%)        & 61.85 & 53.66 & 74.77 & 72.95 & 72.56 & 67.16 \\ \hline
\end{tabular}%
}\vspace{-5mm}
\end{table}

At the post-labeling stage, we compare the proposed ECM with five baselines: % to refine the crowdsourced MSR candidates. 

\begin{itemize}
    \item Random: randomly pick a transcription as the final output
    \item Longest: pick the transcription with the longest text length
    \item Best worker: pick the transcription from the best worker
    \item Oracle: pick the transcription with the lowest edit distance from the ground truth reference
    \item RescoreBERT \cite{9747118}: a BERT-based second-pass rescoring model for ASR error correction. We directly use it to rerank MSR candidates by equally setting the first pass scores to 0.5.
\end{itemize}

ECM is trained on \textit{train-other-10h}. We consider both word frequency and its confidence score, which detects acoustic and semantic defects based on the alignment of audio and text. The optimal trade-off parameter is learnt at 0.8. Table \ref{tab:3} shows our model outperforms the baselines and RescoreBERT by over 30\%. Note that the voting in our method is at word level while RescoreBERT is reranking the entire sentence, so ideally our method may exceed the performance of Oracle.

\subsection{Transcription Error Impact on ASR Model Training}

This section investigates the impact of transcription error on ASR model training. We fine-tune two pre-trained ASR models Wav2Vec2 and WavLM on noisy data, and evaluate their performance on test sets. For fine-tuning, we use the same setup as in \cite{NEURIPS2020_92d1e1eb,9814838}. The ASR models are decoded by a 4-gram LM with decoding parameters tuned on the noisy dev set. Table \ref{tab:5} shows the result of Wav2Vec2 trained on \textit{train-mixed-10h}. Our quality improvement method provides a relative WER reduction of 9.52\% for the ASR system. (R\_WER: 14.49\% → 4.97\%)

% Table 5: Impact of Training Label Quality on ASR model performance (WavLM results listed in Appendix)
\begin{table}[]
\caption{Impact of training data quality on ASR performance}
\vspace{-2mm}
\label{tab:5}
\resizebox{\columnwidth}{!}{%
\begin{tabular}{c|l|ccccc}
\hline
 &
  \multicolumn{1}{c|}{label} &
  \multicolumn{1}{c}{TWER} &
  \multicolumn{1}{c}{\begin{tabular}[c]{@{}c@{}}WER \\ (w/o LM)\end{tabular}} &
  \multicolumn{1}{c}{\begin{tabular}[c]{@{}c@{}}R\_WER \\ (w/o LM)\end{tabular}} &
  \multicolumn{1}{c}{\begin{tabular}[c]{@{}c@{}}WER \\ (w/ LM)\end{tabular}} &
  \multicolumn{1}{c}{\begin{tabular}[c]{@{}c@{}}R\_WER \\ (w/ LM)\end{tabular}} \\ \hline
{test-clean} & Raw Transcription & 5.97          & 10.71          & 12.97\%         & 5.53           & 14.49\%         \\
             & CEM + Random      & 5.00          & 10.27          & 8.33\%          & 5.35           & 10.77\%         \\
             & CEM + ECM         & \textbf{2.99} & \textbf{9.69}  & \textbf{2.22\%} & \textbf{5.07}  & \textbf{4.97\%} \\
             & Ground Truth      & 0.00          & 9.48           & 0.00\%          & 4.83           & 0.00\%          \\ \hline
{test-other} & Raw Transcription & 5.97          & 19.03          & 8.37\%          & 12.01          & 12.56\%         \\
             & CEM + Random      & 5.00          & 18.18          & 3.53\%          & 11.64          & 9.09\%          \\
             & CEM + ECM         & \textbf{2.99} & \textbf{17.65} & \textbf{0.51\%} & \textbf{11.01} & \textbf{3.19\%} \\
             & Ground Truth      & 0.00          & 17.56          & 0.00\%          & 10.67          & 0.00\%          \\ \hline
\end{tabular}%
} \vspace{-5mm}
\end{table}

\begin{figure}[h]
  \centering
  \vspace{-2mm}
  \includegraphics[width=0.7\linewidth]{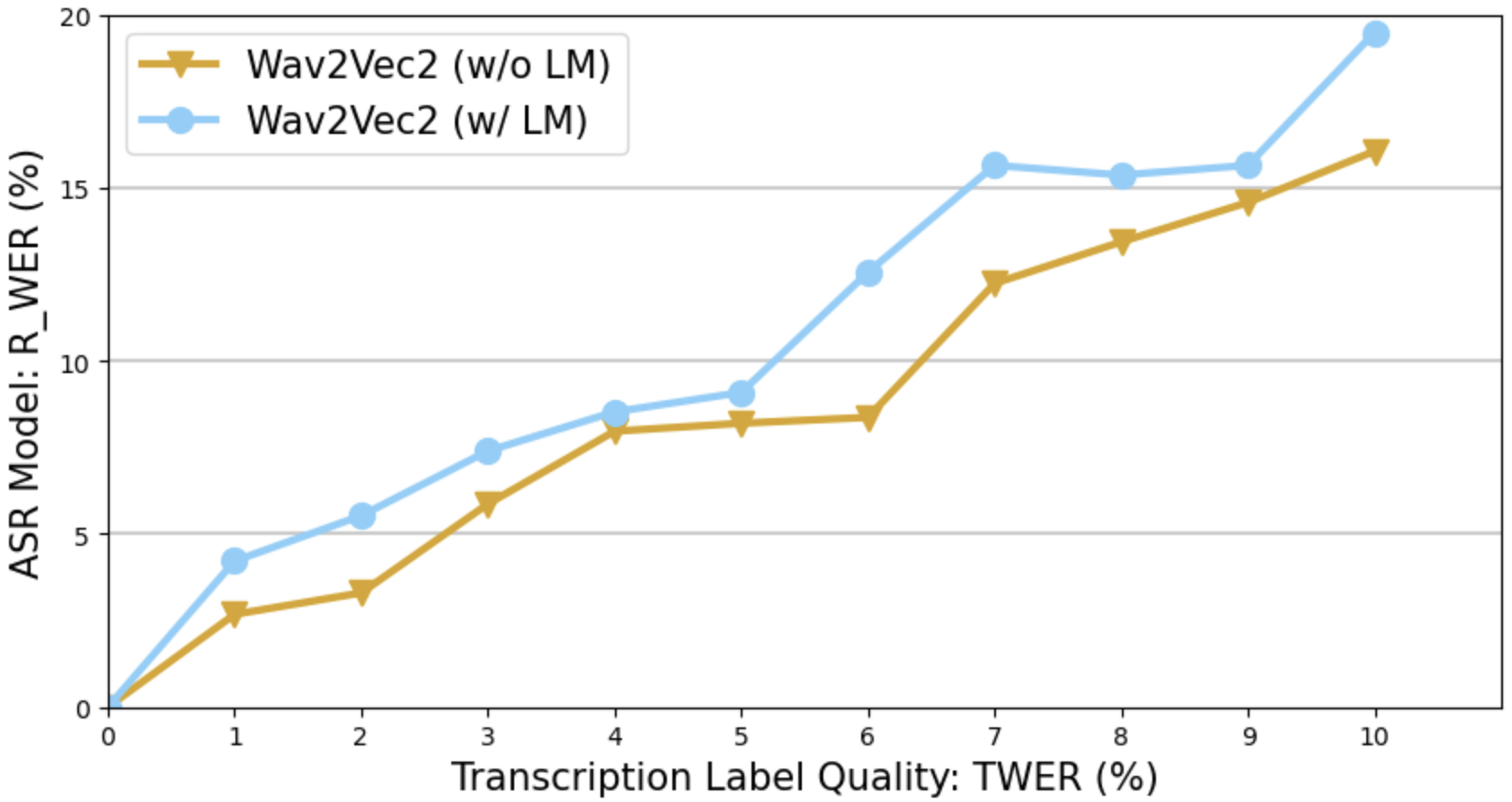}
  \vspace{-2mm}
  \caption{ASR models fine-tuned at controlled noise level}
  \label{fig:twer_wer}
  \vspace{-2.2mm}
\end{figure}

To quantify the relation between training data quality and ASR model's performance, we set TWER at predefined scales (0 - 10\%) by randomly mixing noisy human transcriptions with ground truth reference. We do not use any data augmentation strategy like SpecAugment \cite{park19e_interspeech}. Figure \ref{fig:twer_wer} shows a strong correlation between TWER and R\_WER.

\begin{figure}[h]
  \centering
  \vspace{-2mm}
  \includegraphics[width=0.7\linewidth]{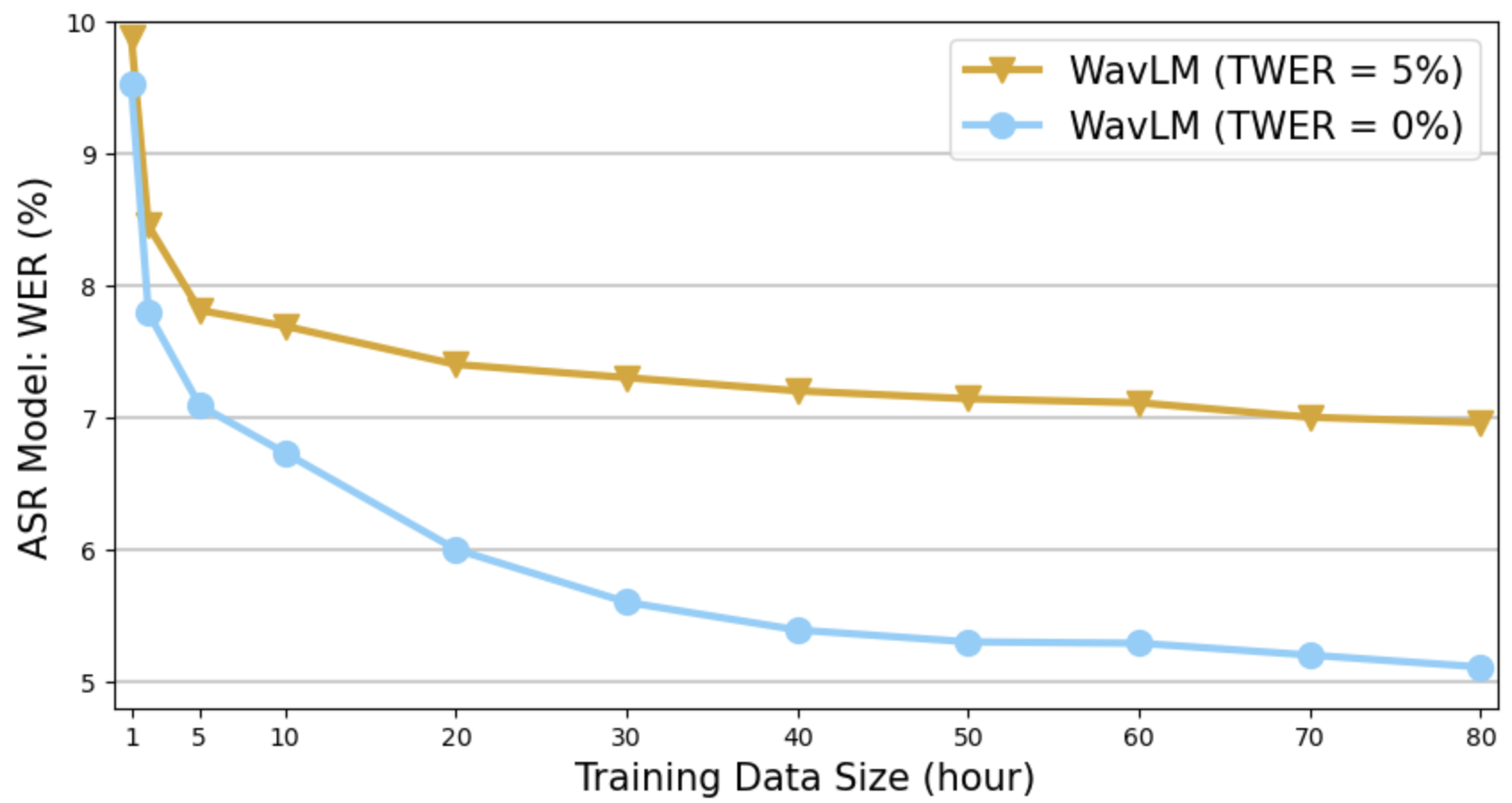}
  \vspace{-2mm}
  \caption{ASR models fine-tuned at controlled training data size}
  \label{fig:datasize}
  \vspace{-3mm}
\end{figure}

We further investigate whether the impact of transcription error can be mitigated by switching to a larger model with more training data. We fine-tune WavLM on clean and noisy data (TWER = 5\%) with training data size increased from 1 to 80 hours. The asymptote gap in Figure \ref{fig:datasize} indicates increasing data and model size cannot mitigate the harm of transcription defect.

\section{Conclusion and Future Work}

In this paper, we propose a new speech transcription data collection method with ML-in-the-loop quality improvement mechanisms to detect and correct human errors. In particular, CEM detects low-quality transcriptions for relabeling, and ECM automatically corrects word errors and generates the final transcription. We investigate the impact of transcription error on ASR model training, and find a strong correlation between label quality and ASR system's performance. 

We collect and release LibriCrowd, a large-scale crowdsourced dataset of human speech transcription. We apply the proposed quality improvement method on the collected human transcriptions. Our method reduces TWER by over 50\%, which gives 10\% relative WER reduction to the ASR models trained on it. We believe this dataset can benefit the research community to develop advanced MSR error correction models, as well as to design ASR algorithms that are robust on noisy data. 

For future work, we will continue to collect crowdsourced data from scripted and unscripted speech to enhance LibriCrowd. We will explore more transcription error mitigation methods such as including large language models.

\bibliographystyle{IEEEtran}
\bibliography{mybib}

\begin{thebibliography}{10}
\providecommand{\url}[1]{#1}
\csname url@samestyle\endcsname
\providecommand{\newblock}{\relax}
\providecommand{\bibinfo}[2]{#2}
\providecommand{\BIBentrySTDinterwordspacing}{\spaceskip=0pt\relax}
\providecommand{\BIBentryALTinterwordstretchfactor}{4}
\providecommand{\BIBentryALTinterwordspacing}{\spaceskip=\fontdimen2\font plus
\BIBentryALTinterwordstretchfactor\fontdimen3\font minus
  \fontdimen4\font\relax}
\providecommand{\BIBforeignlanguage}[2]{{%
\expandafter\ifx\csname l@#1\endcsname\relax
\typeout{** WARNING: IEEEtran.bst: No hyphenation pattern has been}%
\typeout{** loaded for the language `#1'. Using the pattern for}%
\typeout{** the default language instead.}%
\else
\language=\csname l@#1\endcsname
\fi
#2}}
\providecommand{\BIBdecl}{\relax}
\BIBdecl

\bibitem{8049322}
W.~Xiong, J.~Droppo, X.~Huang, F.~Seide, M.~L. Seltzer, A.~Stolcke, D.~Yu, and
  G.~Zweig, ``Toward human parity in conversational speech recognition,''
  \emph{IEEE/ACM Transactions on Audio, Speech, and Language Processing},
  vol.~25, no.~12, pp. 2410--2423, 2017.

\bibitem{10.5555/562393}
H.~A. Bourlard and N.~Morgan, \emph{Connectionist Speech Recognition: A Hybrid
  Approach}.\hskip 1em plus 0.5em minus 0.4em\relax Kluwer Academic Publishers,
  1993.

\bibitem{TIMIT1993}
J.~S. Garofolo and et~al., ``Timit acoustic-phonetic continuous speech
  corpus,'' \emph{Philadelphia: Linguistic Data Consortium}, vol. LDC93S1,
  1993.

\bibitem{mcauliffe17_interspeech}
M.~McAuliffe, M.~Socolof, S.~Mihuc, M.~Wagner, and M.~Sonderegger, ``{Montreal
  Forced Aligner: Trainable Text-Speech Alignment Using Kaldi},'' in
  \emph{Proc. Interspeech 2017}, 2017, pp. 498--502.

\bibitem{Graves:06icml}
A.~Graves, S.~Fernandez, F.~Gomez, and J.~Schmidhuber, ``Connectionist temporal
  classification: Labelling unsegmented sequence data with recurrent neural
  nets,'' in \emph{ICML}, 2006.

\bibitem{DBLP:journals/corr/abs-1211-3711}
A.~Graves, ``Sequence transduction with recurrent neural networks,''
  \emph{CoRR}, vol. abs/1211.3711, 2012.

\bibitem{NEURIPS2020_1457c0d6}
T.~Brown and et~al., ``Language models are few-shot learners,'' in
  \emph{Advances in Neural Information Processing Systems}, H.~Larochelle,
  M.~Ranzato, R.~Hadsell, M.~Balcan, and H.~Lin, Eds., vol.~33.\hskip 1em plus
  0.5em minus 0.4em\relax Curran Associates, Inc., 2020, pp. 1877--1901.

\bibitem{NEURIPS2020_92d1e1eb}
A.~Baevski, Y.~Zhou, A.~Mohamed, and M.~Auli, ``wav2vec 2.0: A framework for
  self-supervised learning of speech representations,'' in \emph{Advances in
  Neural Information Processing Systems}, H.~Larochelle, M.~Ranzato,
  R.~Hadsell, M.~Balcan, and H.~Lin, Eds., vol.~33.\hskip 1em plus 0.5em minus
  0.4em\relax Curran Associates, Inc., 2020, pp. 12\,449--12\,460.

\bibitem{9688253}
Y.-A. Chung, Y.~Zhang, W.~Han, C.-C. Chiu, J.~Qin, R.~Pang, and Y.~Wu,
  ``w2v-bert: Combining contrastive learning and masked language modeling for
  self-supervised speech pre-training,'' in \emph{ASRU}, 2021, pp. 244--250.

\bibitem{9585401}
W.-N. Hsu, B.~Bolte, Y.-H.~H. Tsai, K.~Lakhotia, R.~Salakhutdinov, and
  A.~Mohamed, ``Hubert: Self-supervised speech representation learning by
  masked prediction of hidden units,'' \emph{IEEE/ACM Transactions on Audio,
  Speech, and Language Processing}, vol.~29, pp. 3451--3460, 2021.

\bibitem{9814838}
S.~Chen and et~al., ``Wavlm: Large-scale self-supervised pre-training for full
  stack speech processing,'' \emph{IEEE Journal of Selected Topics in Signal
  Processing}, vol.~16, no.~6, pp. 1505--1518, 2022.

\bibitem{9053896}
Q.~Zhang, H.~Lu, H.~Sak, A.~Tripathi, E.~McDermott, S.~Koo, and S.~Kumar,
  ``Transformer transducer: A streamable speech recognition model with
  transformer encoders and rnn-t loss,'' in \emph{ICASSP}, 2020, pp.
  7829--7833.

\bibitem{gulati20_interspeech}
A.~Gulati, J.~Qin, C.-C. Chiu, N.~Parmar, Y.~Zhang, J.~Yu, W.~Han, S.~Wang,
  Z.~Zhang, Y.~Wu, and R.~Pang, ``{Conformer: Convolution-augmented Transformer
  for Speech Recognition},'' in \emph{Proc. Interspeech 2020}, 2020, pp.
  5036--5040.

\bibitem{han20_interspeech}
W.~Han, Z.~Zhang, Y.~Zhang, J.~Yu, C.-C. Chiu, J.~Qin, A.~Gulati, R.~Pang, and
  Y.~Wu, ``{ContextNet: Improving Convolutional Neural Networks for Automatic
  Speech Recognition with Global Context},'' in \emph{Proc. Interspeech 2020},
  2020, pp. 3610--3614.

\bibitem{7178964}
V.~Panayotov, G.~Chen, D.~Povey, and S.~Khudanpur, ``Librispeech: An asr corpus
  based on public domain audio books,'' in \emph{ICASSP}, 2015, pp. 5206--5210.

\bibitem{DBLP:journals/corr/abs-2109-01164}
\BIBentryALTinterwordspacing
M.~Liu, C.~Zhang, H.~Xing, C.~Feng, M.~Chen, J.~Bishop, and G.~Ngapo,
  ``Scalable data annotation pipeline for high-quality large speech datasets
  development,'' \emph{CoRR}, vol. abs/2109.01164, 2021. [Online]. Available:
  \url{https://arxiv.org/abs/2109.01164}
\BIBentrySTDinterwordspacing

\bibitem{levit2017dont}
M.~Levit, Y.~Huang, S.~Chang, and Y.~Gong, ``Don't count on asr to transcribe
  for you: Breaking bias with two crowds,'' \emph{Interspeech}, August 2017.

\bibitem{1325949}
R.~Sundaram and J.~Picone, ``Effects on transcription errors on supervised
  learning in speech recognition,'' in \emph{2004 IEEE International Conference
  on Acoustics, Speech, and Signal Processing}, vol.~1, 2004, pp. I--169.

\bibitem{DBLP:journals/corr/abs-2112-00350}
\BIBentryALTinterwordspacing
I.~Chen, B.~King, and J.~Droppo, ``Investigation of training label error impact
  on {RNN-T},'' \emph{CoRR}, vol. abs/2112.00350, 2021. [Online]. Available:
  \url{https://arxiv.org/abs/2112.00350}
\BIBentrySTDinterwordspacing

\bibitem{odell1956profit}
M.~K. Odell, ``The profit in records management,'' \emph{Systems (New York)},
  vol.~20, p.~20, 1956.

\bibitem{DBLP:conf/nips/PavlichenkoSU21}
N.~Pavlichenko, I.~Stelmakh, and D.~Ustalov, ``Crowdspeech and vox {DIY:}
  benchmark dataset for crowdsourced audio transcription,'' in \emph{NeurIPS
  Datasets and Benchmarks}, 2021.

\bibitem{ardila-EtAl:2020:LREC}
R.~Ardila, M.~Branson, K.~Davis, M.~Kohler, J.~Meyer, M.~Henretty, R.~Morais,
  L.~Saunders, F.~Tyers, and G.~Weber, ``Common voice: A
  massively-Â­multilingual speech corpus,'' in \emph{Proceedings of The 12th
  Language Resources and Evaluation Conference}.\hskip 1em plus 0.5em minus
  0.4em\relax Marseille, France: European Language Resources Association, May
  2020, pp. 4218--4222.

\bibitem{5494979}
M.~Marge, S.~Banerjee, and A.~I. Rudnicky, ``Using the amazon mechanical turk
  for transcription of spoken language,'' in \emph{2010 IEEE International
  Conference on Acoustics, Speech and Signal Processing}, 2010, pp. 5270--5273.

\bibitem{novotney-callison-burch-2010-cheap}
S.~Novotney and C.~Callison-Burch, ``Cheap, fast and good enough: Automatic
  speech recognition with non-expert transcription,'' in \emph{NAACL}.\hskip
  1em plus 0.5em minus 0.4em\relax Los Angeles, California: Association for
  Computational Linguistics, Jun. 2010, pp. 207--215.

\bibitem{5947474}
K.~Audhkhasi, P.~Georgiou, and S.~S. Narayanan, ``Accurate transcription of
  broadcast news speech using multiple noisy transcribers and unsupervised
  reliability metrics,'' in \emph{ICASSP}, 2011.

\bibitem{10.1007/978-3-030-60276-5_27}
L.~K{\"u}rzinger, D.~Winkelbauer, L.~Li, T.~Watzel, and G.~Rigoll,
  ``Ctc-segmentation of large corpora for german end-to-end speech
  recognition,'' in \emph{Speech and Computer}.\hskip 1em plus 0.5em minus
  0.4em\relax Springer International Publishing, 2020, pp. 267--278.

\bibitem{9688175}
H.~Futami, H.~Inaguma, M.~Mimura, S.~Sakai, and T.~Kawahara, ``Asr rescoring
  and confidence estimation with electra,'' in \emph{2021 IEEE Automatic Speech
  Recognition and Understanding Workshop (ASRU)}, 2021, pp. 380--387.

\bibitem{kim08e_interspeech}
C.~Kim and R.~M. Stern, ``{Robust signal-to-noise ratio estimation based on
  waveform amplitude distribution analysis},'' in \emph{Proc. Interspeech
  2008}, 2008, pp. 2598--2601.

\bibitem{NIPS2017_6449f44a}
G.~Ke, Q.~Meng, T.~Finley, T.~Wang, W.~Chen, W.~Ma, Q.~Ye, and T.-Y. Liu,
  ``Lightgbm: A highly efficient gradient boosting decision tree,'' in
  \emph{Neural Information Processing Systems}, 2017.

\bibitem{devlin-etal-2019-bert}
J.~Devlin, M.-W. Chang, K.~Lee, and K.~Toutanova, ``{BERT}: Pre-training of
  deep bidirectional transformers for language understanding,'' in
  \emph{NAACL}.\hskip 1em plus 0.5em minus 0.4em\relax Minneapolis, Minnesota:
  Association for Computational Linguistics, Jun. 2019, pp. 4171--4186.

\bibitem{lewis-etal-2020-bart}
M.~Lewis and et~al., ``{BART}: Denoising sequence-to-sequence pre-training for
  natural language generation, translation, and comprehension,'' in
  \emph{Proceedings of the 58th Annual Meeting of the Association for
  Computational Linguistics}.\hskip 1em plus 0.5em minus 0.4em\relax
  Association for Computational Linguistics, Jul. 2020, pp. 7871--7880.

\bibitem{9747118}
L.~Xu, Y.~Gu, J.~Kolehmainen, H.~Khan, A.~Gandhe, A.~Rastrow, A.~Stolcke, and
  I.~Bulyko, ``Rescorebert: Discriminative speech recognition rescoring with
  bert,'' in \emph{ICASSP}, 2022, pp. 6117--6121.

\bibitem{NEEDLEMAN1970443}
S.~B. Needleman and C.~D. Wunsch, ``A general method applicable to the search
  for similarities in the amino acid sequence of two proteins,'' \emph{Journal
  of Molecular Biology}, vol.~48, no.~3, pp. 443--453, 1970.

\bibitem{659110}
J.~Fiscus, ``A post-processing system to yield reduced word error rates:
  Recognizer output voting error reduction (rover),'' in \emph{1997 IEEE
  Workshop on Automatic Speech Recognition and Understanding Proceedings},
  1997, pp. 347--354.

\bibitem{9052942}
J.~Kahn, M.~Rivière, W.~Zheng, E.~Kharitonov, Q.~Xu, P.~Mazaré, J.~Karadayi,
  V.~Liptchinsky, R.~Collobert, C.~Fuegen, T.~Likhomanenko, G.~Synnaeve,
  A.~Joulin, A.~Mohamed, and E.~Dupoux, ``Libri-light: A benchmark for asr with
  limited or no supervision,'' in \emph{ICASSP}, 2020, pp. 7669--7673.

\bibitem{park19e_interspeech}
D.~S. Park, W.~Chan, Y.~Zhang, C.-C. Chiu, B.~Zoph, E.~D. Cubuk, and Q.~V. Le,
  ``{SpecAugment: A Simple Data Augmentation Method for Automatic Speech
  Recognition},'' in \emph{Proc. Interspeech 2019}, 2019, pp. 2613--2617.

\end{thebibliography}

\end{document}